# Adaptive Friction in Deep Learning: Enhancing Optimizers with Sigmoid and Tanh Function

Hongye Zheng*
Chinese University of Hong Kong
HongKong, China

Bingxing Wang
Illinois Institute of Technology
Chicago, USA

Minheng Xiao
Ohio State University
Columbus, USA

Honglin Qin
Stevens Institute of Technology
Hoboken, USA

Zhizhong Wu
University of California, Berkeley
Berkeley, USA

LiangHao Tan
Independent Researcher
San Jose, USA

*Abstract*—Adaptive optimizers are pivotal in guiding the weight updates of deep neural networks, yet they often face challenges such as poor generalization and oscillation issues. To counter these, we introduce sigSignGrad and tanhSignGrad, two novel optimizers that integrate adaptive friction coefficients based on the Sigmoid and Tanh functions, respectively. These algorithms leverage short-term gradient information, a feature overlooked in traditional Adam variants like diffGrad and AngularGrad, to enhance parameter updates and convergence.Our theoretical analysis demonstrates the wide-ranging adjustment capability of the friction coefficient S, which aligns with targeted parameter update strategies and outperforms existing methods in both optimization trajectory smoothness and convergence rate. Extensive experiments on CIFAR-10, CIFAR-100, and Mini-ImageNet datasets using ResNet50 and ViT architectures confirm the superior performance of our proposed optimizers, showcasing improved accuracy and reduced training time. The innovative approach of integrating adaptive friction coefficients as plug-ins into existing optimizers, exemplified by the sigSignAdamW and sigSignAdamP variants, presents a promising strategy for boosting the optimization performance of established algorithms. The findings of this study contribute to the advancement of optimizer design in deep learning.

## I. INTRODUCTION

In the training of deep neural networks, the weight distribution of each layer is an important parameter of the network. Adaptive optimizers, in conjunction with the backpropagation algorithm, guide the update of network weight parameters and enhance network performance, which is an essential part of network training. Adaptive methods converge quickly in the early stages of training, but their generalization performance on the test set is weaker than that of non-adaptive methods, and they suffer from sawtooth-like oscillation problems in the iteration trajectory. Many scholars have proposed various Adam[1] variants based on the aforementioned issues, but most Adam family optimization algorithms ignore the role of short-term gradient information in parameter optimization. diffGrad [2] and AngularGrad [3] propose to adjust the learning rate using short-term gradient information, but both have only verified the effectiveness of their methods in image classification tasks. In addition, the complex angle calculation in the AngularGrad algorithm makes it require more time and computational resources in model training.

Based on the above issues, in order to further optimize the optimization performance of adaptive optimizer algorithms, this paper deeply explores the role of short-term gradient information in parameter updates. The main contributions and innovations are as follows:

(1) This paper specifically implements the adaptive friction coefficient S based on the Sigmoid [4] function and the Tanh [5] function. On the basis of Adam, the adaptive friction coefficient S is introduced, thus proposing two new optimizer algorithms, sigSignGrad and tanhSignGrad. The paper theoretically analyzes the adjustment effect of the friction coefficient S on parameter updates in different regions of the loss function.

(2) This paper verifies the possibility of inserting the proposed friction coefficient S as a plug-in into existing optimizers, attempts to insert the friction coefficient S implemented based on the Sigmoid function into AdamW [6] and AdamP [7], thus proposing two new optimization algorithms, sigSignAdamW and sigSignAdamP, providing a scheme for improving the optimization performance of existing optimizers.

(3) This paper verifies the excellent performance of sigSignGrad and tanhSignGrad through a large number of experiments.

## II. BACKGROUND

In recent years, adaptive optimizers have become integral to the training of deep neural networks, addressing issues related to generalization and oscillation. This section reviews the significant contributions and methodologies relevant to the development and enhancement of these optimizers. In the medical domain, deep learning models have seen significant advancements. Xu et al. emphasized the importance of data preprocessing in improving the performance of deep learning models for medical diagnostics [8]. Similarly, Xiao et al. applied conditional Generative Adversarial Networks (cGAN) to enhance surgical imaging by effectively defogging images, demonstrating the versatility and potential of deep learning techniques in medical imaging [9].

The integration of attention mechanisms in deep learning models has also shown considerable promise. Xiao et al. explored the enhancement of deep learning models through attention mechanisms for mining medical textual data, highlighting the importance of efficient data handling and feature extraction [10]. Zhu et al. introduced the Attention-Unet model, which combines attention mechanisms with U-Net architectures to achieve fast and accurate segmentation in medical imaging, thereby improving model performance and training efficiency [11]. Furthermore, the use of adaptive optimizers in neural networks for survival prediction across diverse cancer types has been a significant area of research. Yan et al. employed neural networks to predict patient survival, illustrating the potential of deep learning models to provide valuable insights in medical prognostics [12]. These studies underscore the broad applicability of deep learning methodologies across various applications, aligning with our research focus on enhancing optimizer performance. Here is an introduction to the mainstream optimization algorithms.

### A. AdamW

AdamW discovered that $l_2$ regularization and weight decay are not equivalent concepts in Adam, and thus proposed to decouple the weights on the basis of Adam. $l_2$ regularization adds the square sum of all model weights as a penalty term to the loss function. The loss function with the $l_2$ regular term is detailed in Equation .

$$\min_w L_2(w) = \min_w f(w) + \frac{\lambda}{2n}\sum_{l=1}^{n} w_i^2$$

$$L'_2(w) = f'(w) + \frac{\lambda}{n}\sum_{i=1}^{n} w_i$$

The estimated gradient needs to add the result of the derivative of this regular term, as detailed in Equation .

$$g_t = \nabla_\theta f_t(\theta_{t-1}) + w_t \theta_{t-1}$$

Weight decay is the process of subtracting a small part of the weight when updating the weights. Taking the original SGD as an example, the parameter update formula after introducing weight decay is shown in Equation .

$$g_t = \nabla_\theta f_t(\theta_{t-1}) + w_t \theta_{t-1}$$

AdamW proposed to decouple the $l_2$ regular term from the gradient in Adam, and directly introduce it into the final parameter update to achieve the effect of weight decay and improve generalization performance. The parameter update formula of AdamW is shown in Equation .

$$\theta_t = \theta_{t-1} - \left(\frac{\alpha \widetilde{m}_t}{\sqrt{v_l} + \epsilon} + \lambda \theta_{t-1}\right)$$

### B. AdamP

AdamP found that in momentum-based gradient descent optimizers, scale invariance can lead to excessive growth in the norm of the weights, causing the premature reduction of effective step size during training, which leads to poorer optimization performance. Therefore, AdamP proposes to remove the radial component of the weight vector (i.e., the component parallel to the weight direction) in momentum-based gradient descent algorithms (such as SGD and Adam) to slow down the decay of the effective update step size. The parameter update process of AdamP is shown in Equations.

$$p_t = \frac{m_t}{\sqrt{v_t}+\epsilon}$$

$$\Pi_\theta(x) := x - (\hat{\theta} \cdot x)\hat{\theta}$$

$$q_t = \begin{cases} \Pi_{\theta_t}(p_t), & \text{if } \cos(\theta_t, \nabla_\theta f(\theta_t)) < \delta/\sqrt{\dim(\theta)} \\ p_t, & \text{otherwise} \end{cases}$$

$$\theta_t = \theta_{t-1} - \alpha q_t$$

### C. DiffGrad

DiffGrad proposes to introduce the difference between the current and the most recent iteration gradients to adjust the parameter updates. It introduces a damping factor, DFC, represented by $\xi_t$ in the algorithm, with a value range of (0.5, 1). When the difference between the two gradients is large, diffGrad considers reducing the damping force and taking a larger update step; when the difference between the two gradients is small, it believes that more damping force is needed, reducing the update step size. The parameter update method of diffGrad is shown in Equations.

$$\text{AbsSig}(x) = \frac{1}{1 + e^{-|x|}}$$

$$\xi_t = \text{AbsSig}(g_{t-1} - g_t)$$

$$\theta_t = \theta_{t-1} - \alpha \xi_t \widehat{m}_t / \left(\sqrt{\widehat{v}_t} + \epsilon\right)$$

### D. AngularGrad

AngularGrad believes that diffGrad does not solve the problem of high gradient variance in the early stages of training, and its convergence curve is jagged and vibrating rather than smooth. Therefore, it proposes to adjust the update step size using the angle change information between two consecutive short-term gradients, which is the first time gradient direction and angle information have been applied in an optimizer. The improvements it made on the basis of Adam are shown in Equations.

$$A_t = \tan^{-1}\left|\frac{(g_t - g_{t-1})}{(1 + g_t * g_{t-1})}\right|$$

$$A_{min} = min(A_{t-1}, A_t)$$

$$\phi_t = \text{Tanh}(|(\angle A_{min})|)\lambda_1 + \lambda_2$$

$$\theta_t = \theta_{t-1} - \alpha \phi_t \frac{\widetilde{m}_t}{\sqrt{v_t}+\epsilon}$$

### III. METHOD

### A. SigSignGrad

We will introduce an adaptive friction coefficient S based on the Sigmoid function into Adam, resulting in a new optimization algorithm, sigSignGrad. Our main focus will be on analyzing the adjustment effect of the adaptive friction coefficient S on parameter updates based on sigSignGrad.

The Sigmoid function maps the input parameter to a range between (0,1). The expression for the friction coefficient $S_t$ based on the Sigmoid function is shown in Equation.

$$S_t = \text{Sigmoid}(\beta_3 g_{t-1} * g_t)$$

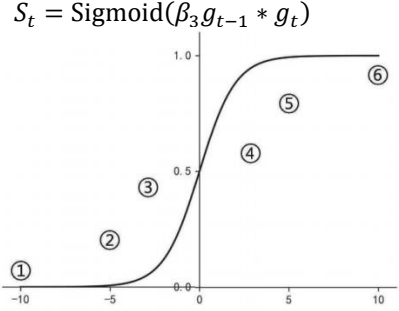

Figure 1. Sigmoid function based on friction coefficient

From Figure 1, it can be observed that when the signs of the two gradients are the same, the friction coefficient S falls within the range (0.5,1). Conversely, when the signs of the two gradients are opposite, the friction coefficient S is within the range (0,0.5), which aligns with the target parameter update strategy. The friction coefficient DFC of diffGrad and the friction coefficient φ of AngularGrad both lie within the range (0.5,1). Compared to diffGrad and AngularGrad, the friction coefficient S based on the Sigmoid function can adjust the parameter update with a broader range of steps, enabling the optimizer to converge more rapidly.

As shown in Figure 2, the black dashed arrows represent the direction of the gradient, where a and b denote two consecutive gradients updated by Adam, with a being the gradient of the most recent iteration and b being the current gradient. $b'$ represents the parameter gradient b after incorporating the friction coefficient S. To simplify the description, we assume that the gradient consists of only two directional components, namely the x-axis and y-axis directions. The friction coefficient S can reduce the update step size of the gradient b in the y-axis direction while increasing the update step size of the gradient b in the x-axis direction. Consequently, the angle between two consecutive gradients changes from the black $A_t$ to the gray $A'_t$, making the optimization trajectory smoother.

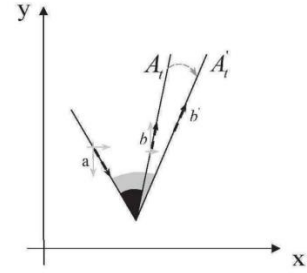

Figure 2. Adjustment effect of frictional coefficient S on gradient updating direction

In the case of the area (c) in Figure 3, after incorporating the friction coefficient S, the angle $A_8$ becomes flatter, and the parameter update from $\theta_8$ to $\theta_9$ will be closer to the local minimum $x^*$.

In summary, the friction coefficient S contributes to smoothing the optimization trajectory, accelerating convergence, and enhancing the optimizer's ability to converge to the minimum value.

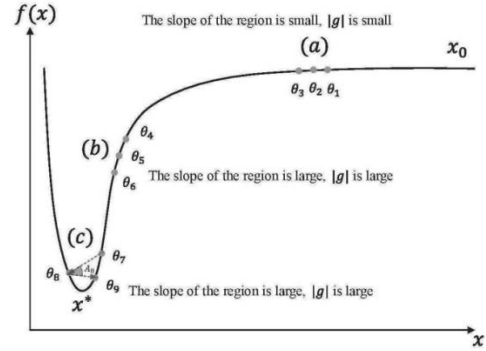

Figure 3. The relationship between curve slope and gradient in different regions of loss function

*B. TanhSignGrad*

We have also considered the feasibility of implementing an adaptive friction coefficient S using the Tanh function, and by introducing this adaptive friction coefficient S based on the Tanh function into Adam, we obtain a new optimizer, which we refer to as tanhSignGrad.

The Tanh function is a common activation function that maps the input parameter to a range between (-1,1). The expression for the friction coefficient St based on the Tanh function is shown in Equation .

$$S_t = \text{Tanh}(\beta_3 g_{t-1} * g_t) + 1.0$$

Consequently, the range of the friction coefficient S is (0,2). When the sign of the current iteration gradient is the same as that of the most recent iteration gradient, the parameters are updated with a larger step size (where S is within the range (1,2)); when the signs of the two gradients are opposite, the parameters are updated with a smaller step size (where S is within the range (0,1)). Similarly, compared to diffGrad and AngularGrad, the adaptive friction coefficient S of tanhSignGrad has a broader range to adjust the learning rate, which is more aligned with the goal of updating parameters with a larger step size to accelerate convergence under appropriate conditions, rather than simply updating with a smaller learning rate than before.

The adjustment effect of the adaptive friction coefficient S on parameter updates based on the Tanh function is similar to that of the friction coefficient S based on the Sigmoid function.

## IV. EXPERIMENT

### A. Experiment Settings

#### 1) Dataset

We selected CIFAR-100 [13], and Mini-ImageNet [14] datasets to conduct image classification experiments using the ResNet50 [15] and ViT [16] networks, comparing the sigSignGrad with currently popular optimizers such as SGDM, Adam, diffGrad, and AngularGrad. At the same time, we tested the performance changes of AdamW and AdamP before and after the introduction of the friction coefficient S based on the Sigmoid function (the new optimizers after the introduction of the friction coefficient S are referred to as sigSignAdamW and sigSignAdamP, respectively), to explore the feasibility of our algorithm as a plugin added to other existing optimizers.

#### 2) Model

ResNet50 and ViT represent two typical network architecture types in image classification tasks[17-18]. ResNet50 is a representative of CNNs, while ViT does not rely on CNNs and is fully based on the Transformer architecture, showing excellent performance in image classification tasks. For ViT, which requires pre-training on large datasets to exhibit outstanding performance, we choose to train from scratch. Although the experimental results are lower than those starting from pre-trained models by Lee et al. [19], our experimental purpose is to compare performance with other excellent optimizers, rather than pursuing the highest accuracy. The experiment refers to the parameter settings of Roy et al. [20].

#### 3) Parameter

For SGDM, we set the initial learning rate to 0.1 and the momentum to 0.9. For other optimizers (including Adam, diffGrad, AngularGrad, AdamW, AdamP, sigSignAdamW, sigSignAdamP, and sigSignGrad), the initial learning rate is set to 0.001. The decay coefficients for the first and second moments of the gradients in Adam and its variants $(\beta_1, \beta_2)$ are set to 0.9 and 0.999, respectively, and the constant $\epsilon$ is set to $1 \times 10^{-8}$ Except for AdamW and sigSignAdamW, the weight decay rate for all optimizers is set to 0.

On the CIFAR datasets, the batch size for all optimizers is 128; on the Mini-ImageNet dataset, for ResNet50, the batch size for the optimizer is 64, and for ViT, the batch size is 128.

### B. Experiment Results Analysis

We conducted comparative experiments on the CIFAR-10, CIFAR-100, and Mini-ImageNet datasets using the ResNet50 and ViT networks to evaluate the optimization performance and training time of SGDM, Adam, diffGrad, AngularGrad, and sigSignGrad. We assessed the performance of the optimizers on the test set using the highest classification accuracy (Top-1 Accuracy) and reported the training time on the training set for the ResNet50 and ViT networks using different optimization algorithms.

Under the same experimental setup, we tested and compared the optimization performance and training time changes between sigSignAdamW and sigSignAdamP with the original AdamW and AdamP.

#### 1) ResNet50

Table 1 ResNet50 Experiment Result Acc

| Algorithm | CIFAR | ImageNet |
| --- | --- | --- |
| SGDM | 69.92±2.30 | 64.98±1.24 |
| Adam | 73.77±0.83 | 65.52±1.16 |
| diffGrad | 75.42±0.09 | 67.99±0.15 |
| AngularGrad | 75.67±0.24 | 68.07±0.36 |
| sigSignGrad | 75.92±0.12 | 68.14±0.14 |

Table 1 reports the highest classification accuracy on the test sets after training on the CIFAR-10, CIFAR-100, and Mini-ImageNet training sets with ResNet50, and also shows the training time of ResNet50 on the CIFAR-100 and Mini-ImageNet training sets.

Observing Table 1, on the CIFAR-10 test set, the accuracy of sigSignGrad is comparable to AngularGrad, about 3.55% higher than SGDM, about 0.39% higher than Adam, and about 0.19% higher than diffGrad. On the CIFAR-100 test set, sigSignGrad achieved the highest accuracy, about 0.25% to 6.00% higher than other optimizers. The results on the Mini-ImageNet test set are similar to those on CIFAR, with sigSignGrad showing the best optimization performance, about 0.07% to 3.16% higher than other optimizers.

#### 2) VIT

We reported the experimental results for the ViT network in Table 2 illustrate the changes in Top-1 accuracy on the test sets after training the ViT network from scratch on the CIFAR-10, CIFAR-100, and MiniImageNet training sets. Table 2 shows the highest classification accuracy of ViT on the test sets of the three datasets, and also reports the training time of ViT on the CIFAR-100 and Mini-ImageNet training sets. From Table 2, it can be seen that on the CIFAR-10, CIFAR-100, and MiniImageNet test sets, sigSignGrad demonstrated the highest accuracy apart from SGDM, about 1.23% to 3.90% higher than Adam, about 0.47% to 1.01% higher than diffGrad, and about 0.50% to 0.52% higher than AngularGrad.

Table 2 VIT Experiment Result Acc

| Algorithm | CIFAR | ImageNet |
| --- | --- | --- |
| SGDM | 56.16±0.39 | 38.69±0.84 |
| Adam | 49.39±0.20 | 33.53±1.42 |

| Algorithm | CIFAR | ImageNet |
|---|---|---|
| diffGrad | 51.36±0.25 | 36.42±0.70 |
| AngularGrad | 51.31±0.24 | 36.93±1.30 |
| sigSignGrad | 51.83±0.35 | 37.43±1.27 |

## V. CONCLUSIONS

To alleviate the oscillation of adaptive methods and address the long training time caused by the complexity of the AngularGrad algorithm, this paper proposes an adaptive friction coefficient S based on the change in the short-term gradient sign and realizes the friction coefficient S based on the Sigmoid function and the Tanh function, thus proposing two different optimization algorithms. Theoretical analysis and extensive experiments have verified the excellent performance of the two algorithms proposed in this paper. The experimental results show that, compared with Adam and existing optimizers that utilize short-term gradients, there is an improvement in performance. The findings of this study have significant implications for various application areas that rely on deep learning[21-22]. In particular, fields such as image recognition[23-24], natural language processing[25], and autonomous systems will benefit from the improved optimization performance. The enhanced convergence rates and accuracy of these new optimizers can lead to more efficient training processes and better model performance, making them valuable tools for researchers and practitioners in these domains. Future work will focus on further refining these optimizers and exploring their applicability across various deep learning tasks and architectures.